\documentclass{article}
\usepackage{ijcai18}

\usepackage{times}
\usepackage{soul}
\usepackage{url}
\usepackage[hidelinks]{hyperref}
\usepackage[utf8]{inputenc}
\usepackage[small]{caption}
\usepackage{amsfonts}
\usepackage{amsmath}
\usepackage{array}
\usepackage{subcaption}
\usepackage{csquotes}
\usepackage{graphicx} 

\title{Contrastive explanations for reinforcement learning in terms of expected consequences}

\author{
J. van der Waa $^1$,
J. van Diggelen$^1$,
K. van den Bosch$^1$,
M. Neerincx$^1$
\\ 
$^1$ Netherlands Organisation for Applied Scientific Research\\
jasper.vanderwaa@tno.nl
}
\begin{document}

\maketitle

\begin{abstract}
Machine Learning models become increasingly proficient in complex tasks. However, even for experts in the field, it can be difficult to understand what the model learned. This hampers trust and acceptance, and it obstructs the possibility to correct the model. There is therefore a need for transparency of machine learning models. The development of transparent classification models has received much attention, but there are few developments for achieving transparent Reinforcement Learning (RL) models. In this study we propose a method that enables a RL agent to explain its behavior in terms of the expected consequences of state transitions and outcomes. First, we define a translation of states and actions to a description that is easier to understand for human users. Second, we developed a procedure that enables the agent to obtain the consequences of a single action, as well as its entire policy. The method calculates contrasts between the consequences of the user’s query-derived policy, and of the learned policy of the agent. Third, a format for generating explanations was constructed. A pilot survey study was conducted to explore preferences of users for different explanation properties. Results indicate that human users tend to favor explanations about policy rather than about single actions.
\end{abstract}

\section{Introduction}
Complex machine learning (ML) models such as Deep Neural Networks (DNNs) and Support Vector Machines (SVMs) perform very well on a wide range of  tasks \cite{lundberg2016unexpected}, but their outcomes are often are often difficult to understand by humans  \cite{weller2017challenges}. Moreover, machine learning models cannot explain how they achieved their results.  Even for experts in the field, it can be very difficult to understand what the model actually learned \cite{samek2016interpreting}. To remedy this issue, the field of eXplainable Artificial Intelligence (XAI) studies how such complex but useful models can be made more understandable \cite{gunning2017explainable}.

Achieving transparency of ML models has multiple advantages \cite{weller2017challenges}. For example, if a model designer knows why a model performs badly on some data, he or she can start a more informed process of resolving the performance issues \cite{kulesza2015principles,papernot2018deep}. However, even if a model has high performance, the users (typically non-experts in ML)  would still like to know why it came to a certain output \cite{miller2017explanation}. Especially in high-risk domains such as defense and health care, inappropriate trust in the output may cause substantial risks and problems \cite{lipton2016mythos,ribeiro2016should}. If a ML model fails to provide transparency, the user cannot safely rely on its outcomes, which hampers the model's applicability \cite{lipton2016mythos}. If, however, a ML-model is able to explain its workings and outcomes satisfactorily to the user, then this would not only improve the user's trust; it would also be able to provide new insights to the user.

For the problem of classification, recent research has developed  a number of promising methods that enable classification models to explain their output \cite{guidotti2018survey}. Several of these methods prove to be model-independent in some way, allowing them to be applied on any existing ML classification model. However, for Reinforcement Learning (RL) models,  there are relatively few methods available \cite{verma2018programmatically,shu2017hierarchical,hein2017interpretable}. The scarcity of methods that enable RL agents to explain their actions and policies towards humans severely hampers the practical applications of RL-models in this field. It also diminishes the, often highly rated, value of RL to Artificial Intelligence \cite{hein2017interpretable,gosavi2009reinforcement}. Take for example a simple agent within a grid world that needs to reach a goal position while evading another actor who could cause termination as well as evading other static terminal states. The RL agent cannot easily explain why it takes the route it has learned as it only knows numerical rewards and its coordinates in the grid. The agent has no grounded knowledge about the 'evil actor' that tries to prevent it from reaching its goal nor has it knowledge of how certain actions will effect such grounded concepts. These state features and rewards are what drives the agent but do not lend themselves well for an explanations as they may not be grounded concepts nor do they offer a reason why the agent behaves a certain way.

Important pioneering work has been done by Hayes and Shah \cite{hayes2017improving}. They developed a  method for eXplainable Reinforcement Learning (XRL) that can generate explanations about a  learned policy in a way that is understandable to humans. Their method converts feature vectors to a list of predicates by using a set of binary classification models. This list of predicates is searched to find sub-sets that tend to co-occur with specific actions. The method provides information about which actions are performed when which state predicates are true. A method that uses the co-occurrence to generate explanations may be useful for small problems, but becomes less comprehensible in larger planning and control problems, because the overview of predicate and action combinations becomes too large. Also, the method addresses only \textit{what} the agent does, and not \textit{why} it acts as it does. In other words, the method presents the user with the correlations between states and the policy but it does not provide a motivation why that policy is used in terms of rewards, or state transitions. 

This study proposes an approach to XRL that allows an agent to answer questions about its actions and policy in terms of their consequences. Other questions unique to RL are also possible, for example those that ask about the time it takes to obtain some goal or those about RL specific problems (loop behavior, lack of exploration or exploitation, etc.). However we believe that a non-expert in RL is mostly interested in the expected consequences of the agent's learned behavior and whether the agent finds these consequences good or bad. This information can be used as an argument why the agent behaves in some way. This would allow human users to gain insight in what information the agent can perceive from a state and which outcomes it expects from an action or state visit. Furthermore, to limit the amount of information of all consequences, our proposed method aims to support contrastive explanations \cite{miller2017explanation}. Contrastive explanations are a way of answering causal 'why'-questions. In such questions, two potential items, the fact and foil, are compared to each other in terms of their causal effects on the world. Contrastive questions come natural between humans and offer an intuitive way of gathering motivations about why one performs a certain action instead of another \cite{miller2017explanation}. In our case we allow the user to formulate a question of why the learned policy $\pi_t$ (the 'fact') is used instead of some other policy $\pi_f$ (the 'foil) that is of interest to the user. Furthermore, our proposed method translates the set of states and actions in a set of more descriptive state classes $\mathbf{C}$ and action outcomes $\mathbf{O}$ similar to that of \cite{hayes2017improving}. This allows the user to query the agent in a more natural way as well as receive more informative explanations as both refer to the same concepts instead of plain features. The translation of state features to more high-level concepts and actions in specific states to outcomes, is also done in the proposed algorithm of \cite{sherstov2005improving}. The translation in this algorithm was used to facilitate transfer learning within a single action over multiple tasks and domains. In our method we used it to create a user-interpretable variant of the underlying Markov Decision Problem (MDP).

For the purpose of implementation and evaluation of our proposed method, we performed a pilot study. In this study, a number of explanation examples were presented to participants to see which of their varying properties are preferred the most. One of the properties was to see whether the participants prefer explanations about the expected consequences of a single-action or the entire policy.

\section{Approach for consequence-based explanations}
\label{sec:abstraction_approach}
The underlying Markov Decision Problem (MDP) of a RL agent consists of the tuple $\langle S,A,R,T, \lambda \rangle$. Here, $S$ and $A$ are the set of states (described by a feature vector) and actions respectively, $R:S \times A \rightarrow \mathbb{R}$ is the reward function and $T:S \times A \rightarrow Pr(S)$ the transition function that provides a probability distribution over states. Also, $\lambda$ is the discount factor that governs how much of future rewards are taken into account by the agent. This tuple provides the required information to derive the consequences of the learned policy $\pi_t$ or the foil policy $\pi_f$ from the user's question. As one can use the transition function $T$ to sample the effects of both $\pi_t$ and $\pi_f$. In the case $T$ is not explicit, one may use a separate ML model to learn it in addition to the actual agent. Through this simulation, one constructs a Markov Chain of state visits under each policy $\pi_t$ and $\pi_f$ and can present the difference to the user.

Through the simulation of future states with $T$, information can be gathered about state consequences. In turn, from the agent itself the state or state-action values for simulated state visits can be obtained to develop an explanation in terms of rewards. However, the issue with this approach is that the state features and rewards may not be easy to understand for a user as it would consist of possibly low-level concepts and numerical reward values or expected returns. To mitigate this issue we can apply a translation of the states and actions to a set of predefined state concepts and outcomes. These concepts can be designed to be more descriptive and informative for the potential user. A way to do this translation is by training a set of binary classifiers to recognize each outcome or state concept from the state features and taken action, a similar approach to the one from \cite{hayes2017improving}. Their training can occur during the exploratory learning process of the agent. This translation allows us to use the above described method of simulating consequences and transform the state features and results of actions to more user-interpretable concepts.

\subsection{A user-interpretable MDP}
The original set of states can be transformed to a more descriptive set $\mathbf{C}$ according to the function $\mathbf{k}: S \rightarrow \mathbf{C}$. This is similar to the approach of \cite{hayes2017improving} where $\mathbf{k}$ consists of a number of classifiers. Also, rewards can be explained in terms of a set of action outcomes $\mathbf{O}$ according to $\mathbf{t} : \mathbf{C} \times A \rightarrow Pr(\mathbf{O})$. This provides the results of an action in some state in terms of the concepts $\mathbf{O}$. For example, the outcomes that the developer had in mind when designing the reward function $R$. The transformation of states and actions in state classes and outcomes is adopted from the work of \cite{sherstov2005improving} where the transformations are used to allow for transfer learning in RL. Here however, we use them as a translation towards a more user-interpretable representation of the actual MDP.

\begin{figure}
    \centering
    \includegraphics[width=1.0\columnwidth]{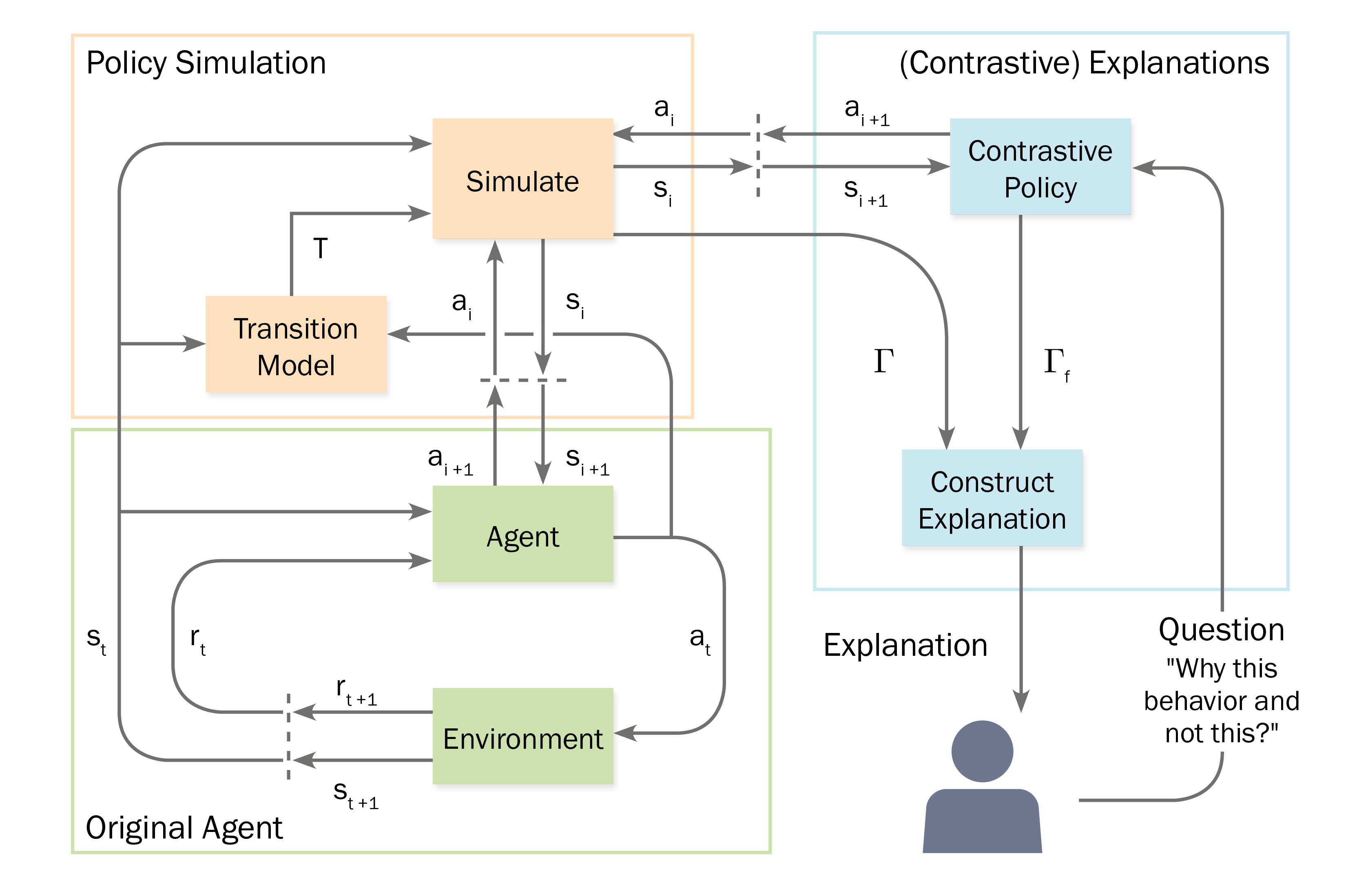}
    \caption{An overview of the proposed method, a dotted line represents a feedback loop. We assume a general reinforcement learning agent that acts upon a state $s$ through some action $a$ and receives a reward $r$. We train a transition model $T$ that can be used to simulate the effect of actions on states. By repeatedly simulating a state $s_i$ we can obtain the expected consequences $\gamma$ of an entire policy. Also, the consequences of a contrastive policy consisting of an alternative courses of action $a_f$ can be simulated with the same transition model $T$. Finally, in constructing the explanation we transform states and actions into user-interpretable concepts and construct an explanation that is contrastive.}
    \label{fig:overview}
\end{figure}

The result is the new MDP tuple $\langle S,A,R,T, \lambda,\mathbf{C},\mathbf{O},\mathbf{t},\mathbf{k} \rangle$. An RL agent is still  trained on $S$, $A$, $R$ and $T$ with $\lambda$ independent of the descriptive sets $\mathbf{C}$ and $\mathbf{O}$ and functions $\mathbf{k}$ and $\mathbf{t}$. This makes the transformation independent of the RL algorithm used to train the agent. See Figure \ref{fig:overview} for an overview of this approach.

\begin{figure}
    \centering
    \includegraphics[width=1.0\columnwidth]{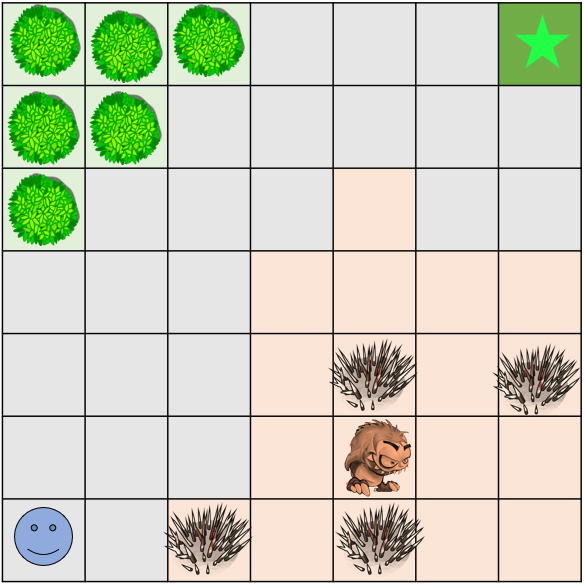}
    \caption{A simple RL problem where the agent has to navigate from the bottom left to the top right (goal) while evading traps, a monster and a forest. The agent terminates when in a tile with a trap or adjacent to the monster. The traps and the monster only occur in the red-shaded area and as soon as the agents enter this area the monster moves towards the agent.}
    \label{fig:simple_case}
\end{figure}

As an example take the grid world illustrated in Figure \ref{fig:simple_case} that shows an agent in a simple myopic navigation task. The states $S$ are the $(x,y)$ coordinates and the presence of a forest, monster or trap in adjacent tiles with $A={Up, Down, Left, Right}$. $R$ consists of a small transient penalty, a slightly larger penalty for tiles with a forest, a large penalty shared over all terminal states (traps or adjacent tiles to a monster) and a large positive reward for the finishing state. $T$ is skewed towards the intended result with small probabilities for the other results if possible.

The state transformation $\mathbf{k}$ can consist out of a set of classifiers for the predicates whether the agent is next to a forest, a wall, a trap or monster, or in the forest. Applying $\mathbf{k}$ to some state $s\in S$ results in a Boolean vector $c\in C$ whose information can be used to construct an explanation in terms of the stated predicates. The similar outcome transformation $\mathbf{t}$ may predict the probability of the outcomes $\mathbf{O}$ given a state and action. In our example, $\mathbf{O}$ consists of whether the agent will be at the goal, in a trap, next to the monster or in the forest. Each outcome $\mathbf{o}$ can be flagged as being positive $\mathbf{o}^+$ or negative $\mathbf{o}^-$ purely such that they can be presented differently in the eventual explanation.

Given the above transformations we can simulate the next state of a single action $a$ with $T$ or even the entire chain of actions and visited states given some policy $\pi$. These can then be transformed into  state descriptions $\mathbf{C}$ and action outcomes $\mathbf{O}$ to form the basis of an explanation. As mentioned, humans usually ask for contrastive questions  especially regarding their actions \cite{miller2017explanation}. In the next section we propose a method of translating the foil in a contrastive question into a new policy.

\subsection{Contrastive questions translated into value functions}
\label{sec:contrast_approach}
A contrastive question consists of a fact and a foil, and its answer describes the contrast between the two from the fact's perspective \cite{miller2017explanation}. In our case, the fact consists of the entire learned policy $\pi_t$, a single action from it $a_t=\pi_t(s_t)$ or any number of consecutive actions from $\pi_t$. We propose a method of how one can obtain a foil policy $\pi_f$ based on the foil in the user's question. An example of such a question could be (framed within the case of Figure \ref{fig:simple_case});

\begin{quotation}
"Why do you move up and then right \textit{(fact)} instead of moving to the right until you hit a wall and then move up \textit{(foil)}?"
\end{quotation}

The foil policy $\pi_f$ is ultimately obtained by combining a state-action value function $Q_I$ -- that represents the user's preference for some actions according to his/her question  -- with the learned $Q_t$ to obtain $Q_f$;

\begin{equation}
    Q_f(s,a) = Q_t(s,a) + Q_I(s,a)\textnormal{, } \forall s,a \in S,A
    \label{eq:impose}
\end{equation}
Each state-action value is of the form $Q:S \times A \rightarrow \mathbb{R}$.

$Q_I$ only values the state-action pairs queried by the user. For instance, the $Q_I$ of the above given user question can be based on the following reward scheme for all potentially simulated $s \in S$; 
\begin{itemize}
    \item The action $a_f^1=\textnormal{'Right'}$ receives a reward such that $Q_f(s, \textnormal{Right}) > Q_t(s, \pi_t(s))$
    \item If $\textnormal{'RightWall'} \in \mathbf{k}(s)$
    \item Then the action $a_f^2=\textnormal{'Up'}$ receives a reward such that $Q_f(\cdot, \textnormal{Up}) > Q_t(\cdot, \pi_t(s))$.
\end{itemize}

Given this reward scheme we can train $Q_I$ and obtain $Q_f$ according to equation \ref{eq:impose}. The state-action values $Q_f$ can then be used to obtain the policy $\pi_f$ using the original action selection mechanism of the agent. This results in a policy that tries to follow the queried policy as best as it can. The advantage of having $\pi_f$ constructed from $Q_f$ is that the agent is allowed to learn a different action then those in the user's question as long as the reward is higher in the long run (more user defined actions can be performed). Also, it allows for the simulation of the actual expected behavior of the agent as it is still based on the agent's action selection mechanism. This would both not be the case if we simply forced the agent to do exactly what the user stated.

The construction of $Q_I$ is done through simulation with the help of the transition model $T$. The rewards that are given during the simulation are selected with Equation \ref{eq:impose} in mind, as they need to eventually compensate for the originally learned action based on $Q_t$. Hence, the reward for each state and queried action is as follows;

\begin{equation}
    R_I(s_{i},a_f) = \frac{\lambda_f}{\lambda} w(s_i,s_t) \left[ R(s_{i},a_f) - R(s_{i},a_t \right] (1+\epsilon)
    \label{eq:foil_reward}
\end{equation}

With $a_t=\pi_t(s_t)$ the originally learned action and $w$ being a distance based weight;

\begin{equation}
w(s_i,s_t) = e^{-\left(\frac{d(s_{i},s_t)}{\sigma}\right)^2}
\end{equation}

First, $s_{i}$ with $i \in \{t,t+1, ..., t+n\}$ is the i'th state in the simulation starting with $s_t$. $a_f$ is the current foil action governed by the conveyed policy by the user. The fact that $a_f$ is taken as the only rewarding action each time, greatly reduces the time needed to construct $Q_I$. Next, $w(s_i,s_t)$ is obtained from a Radial Basis Function (RBF) with a Gaussian kernel and distance function $d$. This RBF represents the exponential distance between our actual state $s_t$ and the simulated state $s_i$. The Gaussian kernel is governed by the standard deviation $\sigma$ and allows us to reduce the effects of $Q_I$ as we get further from our actual state $s_t$. The ratio of discount factors $\frac{\lambda_f}{\lambda}$ allows for the compensation between the discount factor $\lambda$ of the original agent and the potentially different factor $\lambda_f$ for $Q_I$ if we wish it to be more shortsighted. Finally, $\left[ R(s_{i},a_f) - R(s_{i},a_t) \right] (1+\epsilon)$ is the amount of reward that $a_f$ needs such that $Q_I(s_{i}, a_f)\epsilon > Q(s_{i},a_t)$. With $\epsilon > 0$ that determines how much more $Q_I$ will prefer $a_f$ over $a_t$.

The parameter $n$ defines how many future state transitions we simulate and are used to retrieve $Q_I$. As a general rule $n\geq3\sigma$ as at this point the Gaussian kernel will reduce the contribution of $Q_I$ to near zero such that $Q_f$ will resemble $Q_t$. Hence, by setting $\sigma$ one can vary the number of states the foil policy should starting from $s_t$. Also, by setting $\epsilon$ the strength of how much each $a_f$ should be preferred over $a_t$ can be regulated. Finally, $\lambda_f$ defines how shortsighted $Q_I$ should be. If set to $\lambda_f=0$, $\pi_f$ will force the agent to perform $a_f$ as long as $s_i$ is not to distant from $s_t$. If set to values near one, $\pi_f$ is allowed to take different actions as long as it results into more possibilities of performing $a_f$.

\subsection{Generating explanations}
\label{sec:generate_approach}
At this point we have the user-interpretable MDP consisting of state concepts $\mathbf{C}$ and action outcomes $\mathbf{O}$ provided by their respective transformation function $\mathbf{k}$ and $\mathbf{t}$. Also, we have a definition of $R_I$ that values the actions and/or states that are of interest by the user which can be used to train $Q_I$ through simulation and obtain $Q_f$ according to Equation \ref{eq:impose}. This provides us with the basis of obtaining the information needed to construct an explanation.

As mentioned before, the explanations are based on simulating the effects with $T$ of $\pi_t$ and that of $\pi_f$ (if defined by the user). We can call $T$ on the previous state $s_{i-1}$ for some action $\pi(s_{i-1}$ to obtain $s_i$ and repeat this until $i==n$. The result is a single sequence or trajectory of visited states and performed actions for any policy $\pi$ starting from $s_t$;

\begin{equation}
    \gamma(s_t, \pi) = \left\{(s_0, a_0),...,(s_n, a_n)\mid T,\pi \right\}
    \label{eq:simulation}
\end{equation}

If $T$ is probabilistic, multiple simulations with the same policy and starting state may result in different trajectories. To obtain the most probable trajectory $\gamma(s_t, \pi)^*$ we can take the transition from $T$ with the highest probability. Otherwise a Markov chain could be constructed instead of a single trajectory.

The next step is to transform each state and action pair in $\gamma(s_t, \pi)^*$ to the user-interpretable description with the functions $\mathbf{k}$ and $\mathbf{t}$;
\begin{multline}
    Path(s_t, \pi) = \left\{(c_0, o_0),...,(c_n, o_n) \right\}\textnormal{, } \\ c_i=\mathbf{k}(s_i)\textnormal{, } o_i=\mathbf{t}(s_i,a_i), (s_i,a_i)\in \gamma(s_t,\pi)^*
    \label{eq:trajectory}
\end{multline}
From $Path(s_t,\pi_t)$ an explanation can be constructed about the state the agent will most likely visit and the action outcomes it will obtain. For example with the use of the following template;

\begin{quotation}
    "For the next $n$ actions I will mostly perform $a$. During these actions, I will come across situations with $\forall c \in Path(s_t,\pi_t)$. This will cause me $\forall o^+ \in Path(s_t,\pi_t)$ but also $\forall o^- \in Path(s_t,\pi_t)$".
\end{quotation}

Let $a$ here be the action most common in $\gamma(s_t,\pi_t)$ and both $o^+$ and $o^-$ the positive and negative action outcomes respectively. Since we have access to the entire simulation of $\pi_f$, a wide variety of explanations is possible. For instance we could also focus on the less common actions;

\begin{quotation}
    "For the next $n$ actions I will perform $a_1$ when in situations with $\forall c \in Path(s_t,\pi_t|\pi_t=a_1)$ and $a_2$ when in situations with $\forall c \in Path(s_t,\pi_t|\pi_t=a_2)$. These actions prevent me from $\forall o^+ \in Path(s_t,\pi_t)$ but also $\forall o^- \in Path(s_t,\pi_t)$".
\end{quotation}

A contrastive explanation given some question from the user that describes the foil policy $\pi_f$ can be constructed in a similar manner but take the contrast. Given a foil we can focus on the differences between $Path(s_t,\pi_t)$ and $Path(s_t,\pi_f)$. This can be obtained by taking the relative complement $Path(s_t,\pi_t) \backslash Path(s_t,\pi_f)$; the set of expected unique consequences when behaving according to $\pi_t$ and not $\pi_f$. A more extensive explanation can be given by taking the symmetric difference $Path(s_t,\pi_t) \triangle Path(s_t,\pi_f)$ to explain the unique differences between both policies.

\section{User study}
\label{sec:user_study}
The above proposed method allows an RL agent to explain and motivate its behavior in terms of expected states and outcomes. It also enables the construction of contrastive explanations where any policy can be compared to the learned policy. This contrastive explanation is based on differences in expected outcomes between the compared policies.

We performed a small user study in which 82 participants were shown a number of exemplar explanations about the case shown in figure \ref{fig:simple_case}. These explanations addressed either the single next action or the policy. Both explanations can be generated by the above method by adjusting the Radial Basis Function weighting scheme and/or the foil's discount factor. Also, some example explanations were contrastive with only the second best action or policy, while others provided all consequences. Contrasts were determined using the relative complement between fact and foil. Whether the learned action or policy was treated as the fact or foil, was also systematically manipulated in this study.

We presented the developed exemplar explanations in pairs to the participants and asked them to select the explanation that helped them most to understand the agent's behavior. Afterwards we asked which of the following properties they used to assess their preference: long versus short explanations; explanations with ample information versus little  information; explanations addressing actions versus those that address strategies (policies); and explanations addressing short-term consequences of actions versus explanations that address distant consequences of actions.

\begin{figure}
    \centering
    \includegraphics[width=0.95\columnwidth]{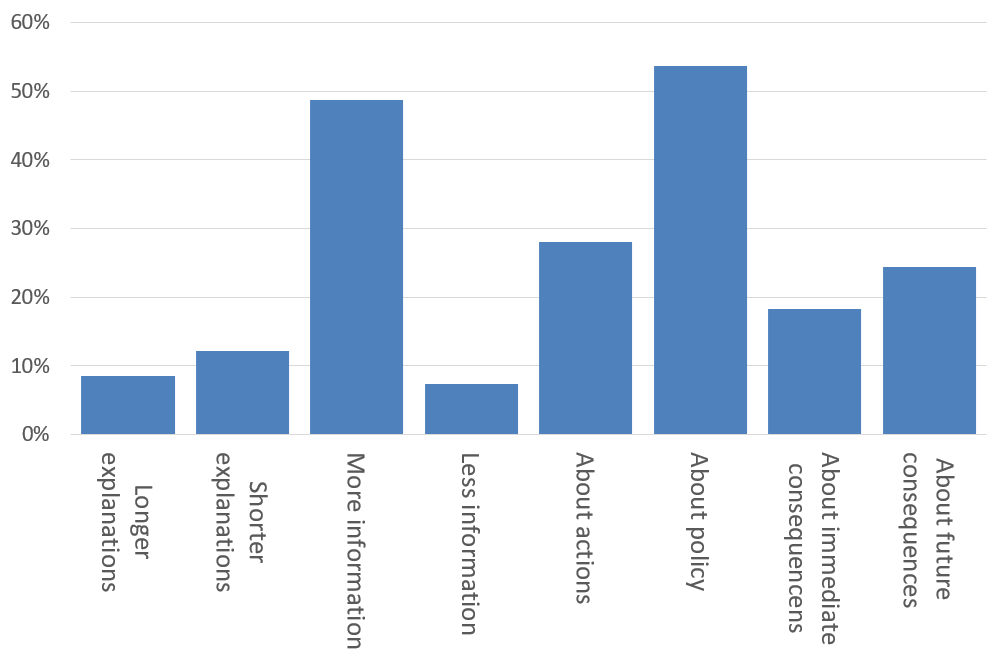}
    \caption{A plot depicting the percentage of participants (y-axis) for each explanation property (x-axis) that caused them to prefer some explanations over others. Answers of a total of 82 participants where gathered.}
    \label{fig:results}
\end{figure}

The results of the preferred factors are shown in Figure \ref{fig:results}. This shows that the participants prefer explanations that address strategy and policy, and that provide ample information. We note here that, given the simple case from figure \ref{fig:simple_case}, participants may have considered an explanation addressing a single action only as trivial, because the optimal action was, in most cases, already evident to the user. 

\section{Conclusion}
We proposed a method for a reinforcement learning (RL) agent to generate  explanations for its actions and strategies. The explanations are based on the expected consequences of its policy. These consequences were obtained through simulation according to a (learned) state transition model. Since state features and numerical rewards do not lend themselves easily for an explanation that is informative to humans, we developed a framework that translates states and actions into user-interpretable concepts and outcomes.

We also proposed a method for converting the foil, --or policy of interest to the user--, of a contrastive 'why'-question about actions into a policy. This policy follows locally the user's query but gradually transgresses back towards the original learned policy. This policy favors the actions that are of interest to the user such that the agent tries to perform them as best as possible. How much these actions are favored compared to the originally learned action can be set with a single parameter.

Through running simulations for a given number steps of both the policy derived from the user's question and the actually learned policy, we were able to obtain expected consequences of each. From here, we were able to construct contrastive explanations: explanations addressing the consequences of the learned policy and what would be different if the derived policy would have been followed.

An online survey pilot study was conducted to explore which of several explanations are most preferred by human users. Results indicate that users prefer explanations about policies rather than about single actions. 

Future work will focus on implementing the method on complex RL benchmarks to explore the scalability of this approach in realistic cases. This is important given the computational costs of simultaneously simulating the consequences of different policies in large state spaces. Also, we will explore more methods to construct our translation functions from states and actions to concepts and outcomes. A more extensive user study will be carried out to evaluate the instructional value of generated explanations in more detail, and to explore the relationship between explanations and users' trust in the agent's performance.

\section*{Acknowledgments}
We would like to thank the reviewers for their time and effort in improving this paper. Also, we are grateful for the funding from the RVO Man Machine Teaming research project that made this research possible.

\bibliographystyle{named}
\bibliography{references}

\end{document}